\documentclass{midl} 

\usepackage{xspace}         
\usepackage{bm}

\usepackage{fancyhdr}       

\usepackage{etoolbox}

\usepackage{appendix}

\usepackage{graphicx}   
\usepackage{graphics}   
\usepackage{wrapfig}    
\usepackage{caption}    

\usepackage{tikz}     
\usetikzlibrary{shapes.geometric, arrows}
\usepackage[framemethod=TikZ]{mdframed}

\usepackage{xifthen}
\usepackage{xargs}

\usepackage[maxlevel=3]{csquotes} 
\usepackage{setspace}             
\usepackage{siunitx} 
\usepackage{amsmath}        
\usepackage{mathrsfs}       
\usepackage{amssymb}        
\usepackage{amsfonts}       
\usepackage{mathtools}      
\usepackage{dsfont}         
\usepackage{stmaryrd}       
\usepackage{esvect}         
\usepackage{systeme}        

\usepackage{tcolorbox}
\usepackage{enumitem}

\usepackage{booktabs}   
\usepackage{pifont}     
\usepackage{makecell}

\usepackage[table]{xcolor}

\newcommand{\braces}[1]{\left\lbrace #1 \right\rbrace} 
\newcommand{\bracks}[1]{\left\lbrack #1 \right\rbrack} 
\newcommand{\pars}[1]{\left( #1 \right)}               

\newcommandx{\proba}[3][1, 3=0]{
\ifthenelse{\isempty{#2}}{\mathbb{P}_{#1}}{%
\ifthenelse{\equal{#3}{0}}{\mathbb{P}_{#1}\pars{#2}}{\mathbb{P}_{#1}\pars{#2 \middle| #3}}}%
}
\newcommandx{\esp}[3][1, 3=0]{
\ifthenelse{\isempty{#2}}{\mathbb{E}_{#1}}{%
\ifthenelse{\equal{#3}{0}}{\mathbb{E}_{#1}\bracks{#2}}{\mathbb{E}_{#1}\bracks{#2 \middle| #3}}}%
}
\newcommandx{\var}[3][1, 3=0]{
\ifthenelse{\isempty{#2}}{\operatorname{Var}_{#1}}{%
\ifthenelse{\equal{#3}{0}}{\operatorname{Var}_{#1}\pars{#2}}{\operatorname{Var}_{#1}\pars{#2 \middle| #3}}}%
}
\newcommandx{\cov}[4][1, 4=0]{
\ifthenelse{\isempty{#2} \AND \isempty{#3}}{\operatorname{Cov}_{#1}}{%
\ifthenelse{\equal{#4}{0}}{\operatorname{Cov}_{#1}\pars{#2, #3}}{\operatorname{Cov}_{#1}\pars{#2, #3 \middle| #4}}}%
}
\newcommandx{\corr}[3][1]{\ifthenelse{\isempty{#2} \AND \isempty{#3}}{\operatorname{Corr}_{#1}}{\operatorname{Corr}_{#1}\pars{#2, #3}}}

\newcommandx{\Norm}[3][1]{\mathcal{N}_{#1}\pars{#2, #3}} 

\newcommandx{\egalloi}{\overset{\mathrm{\mathcal{L}oi}}{=}} 
\newcommandx{\egalprob}{\overset{\mathbb{P}}{=}}            
\newcommandx{\egaltxt}[1]{\overset{\mathrm{#1}}{=}}         
\newcommandx{\simiid}{\overset{\mathrm{iid}}{\sim}}         

\newcommandx{\maxx}[1]{\underset{#1}{\max}}
\newcommandx{\minn}[1]{\underset{#1}{\min}}
\newcommandx{\argmax}[1][1]{\underset{#1}{\arg\max}}
\newcommandx{\argmin}[1][1]{\underset{#1}{\arg\min}}

\newcommandx{\integ}[4][2]{\int_{#1}^{#2} #3 \, \mathrm{d} #4} 

\newcommandx{\surf}[2][2=m]{\numprint{#1}\,#2\textsuperscript{2}}
\newcommandx{\vol}[2][2=m]{\numprint{#1}\,#2\textsuperscript{3}}

\newcommand*\fonction[5]{
#1 \colon \left\{\begin{alignedat}{2}  &#2 &\: &\to      #3\\
                                &#4 &   &\mapsto  #5
\end{alignedat} \right. \kern-\nulldelimiterspace} 

\newcommand{\zs}[1]{\bm z_{s_{#1}}}

\newcommand{\ztsh}[1]{\bm z^{sh}_{t_{#1}}}
\newcommand{\ztsp}[1]{\bm z^{sp}_{t_{#1}}}

\renewcommand{\citeauthor}[1]{\textcolor{blue}{[\citenum{#1}]}}
\renewcommand{\citeyear}[1]{\textcolor{blue}{[\citenum{#1}]}}
\renewcommand{\cite}[1]{\textcolor{blue}{[\citenum{#1}]}}

\newcommand{\bracketcite}[1]{[\citenum{#1}]}

\usepackage{mwe} 
\jmlryear{2025}

\jmlrworkshop{Full Paper -- MIDL 2025}

\jmlrvolume{-- nnn}

\editors{Accepted for publication at MIDL 2025}

\title[DAFTED: Decoupled Asymmetric Fusion of Tabular and Echocardiographic Data]{DAFTED: Decoupled Asymmetric Fusion of Tabular and Echocardiographic Data for Cardiac Hypertension Diagnosis}

\midlauthor{\Name{J\'er\'emie Stym-Popper\nametag{$^{1}$}} \Email{stympopper@isir.upmc.fr}\\
\Name{Nathan Painchaud\nametag{$^{2}$}} \Email{nathan.painchaud@insa-lyon.fr}\\
\Name{Cl\'ement Rambour\nametag{$^{1}$}} \Email{clement.rambour@isir.upmc.fr}\\
\Name{Pierre-Yves Courand\nametag{$^{2}$}} \Email{pierre-yves.courand@chu-lyon.fr}\\
\Name{Nicolas Thome\nametag{$^{1,}\,^{3}$}} \Email{nicolas.thome@isir.upmc.fr} \\
\Name{Olivier Bernard\nametag{$^{2}$}} \Email{olivier.bernard@insa-lyon.fr} \\
\addr $^{1}$ \scriptsize Sorbonne Universit\'e, CNRS, ISIR, MLIA, F-75005 Paris, France \\
\addr $^2$ \scriptsize INSA‐Lyon, Universit\'e Claude Bernard Lyon 1, CNRS, Inserm, CREATIS UMR 5220, U1294, F‐69621, France \\
\addr $^3$ \scriptsize Institut Universaitaire de France (IUF)
}
\def\eg{\textit{e.g.,~}} 
\def\ie{\textit{i.e.,~}}

\begin{document}

\maketitle

\begin{abstract}
Multimodal data fusion has emerged as a key approach in recent years for enhancing diagnosis and prognosis in many medical applications. With the advent of transformer-based methods, it is now possible to combine information from different modalities that provide complementary insights. However, most existing methods rely on symmetric fusion schemes, assuming equal importance for information carried by each modality—a strong assumption that may not always hold true. In this study, we propose an alternative fusion strategy based on an asymmetric scheme. Starting with a primary modality that offers the most critical information, we integrate secondary modality contributions by disentangling shared and modality-specific information. The proposed model was validated on a dataset of 239 patients for characterizing hypertension severity by fusing time series automatically extracted from echocardiographic image sequences and tabular data from patient records. Results show that our approach outperforms existing unimodal and multimodal approaches, achieving an AUC score over 90\% - a crucial benchmark for clinical use.
\end{abstract}

\begin{keywords}
Multimodal fusion, transformers, tables, echocardiography, hypertension.
\end{keywords}

\section{Introduction}

Artificial Intelligence (AI) and deep learning have significantly improved computer-aided diagnosis (CADx) over the last decade \citep{10.1117/12.2254423, 10.3389/fphys.2022.918381,xu_multimodal_2023}. This paper focuses on the characterization of cardiac hypertension (HT). Physicians integrate complementary data from diverse sources, including time-series features derived from echocardiographic sequences and Electronic Health Records (EHRs), to build a comprehensive assessment of the patient's condition \citep{mancia_2023_2023}. Additional measurements, such as 24-hour systolic and diastolic blood pressures (SBP/DBP), are often collected to eliminate ambiguities regarding the severity of the disease. However, this process can be burdensome for patients. In this work, we propose a method to efficiently integrate a tabular representation of minimally invasive EHR data with cardiac time series automatically extracted from apical two and four chamber views (A2C and A4C) using the segmentation framework described in \citep{CARDINAL2023}. This approach aims to enable effective stratification of hypertension while improving patient care.

Combining heterogeneous modalities, such as tabular data and time series, is a non-trivial challenge.
For tabular data, tree-based models, \eg gradient boosting \citep{Chen2016XGboost}, remain the dominant approach \citep{grinsztajn2022why}.
However, a naive fusion strategy that employs XGBoost on both tabular and echocardiographic inputs results in a significant drop in performance compared to using tabular data alone, as observed in related contexts~\citep{wangMediTabScalingMedical2024a}. This can be attributed to the asymmetry of our fusion problem: the primary modality (tabular data) serves as the main source of information, while the secondary modality (time series) provides complementary details but is insufficient for accurate diagnosis on its own. Consequently, there is a need for refined and specialized multimodal fusion methods.

In the recent wave of attentional models and transformers, significant efforts have been devoted to performing multimodal fusion for medical diagnosis.
~The FT-Transformer \citep{FT-gorishny2021, zhu2023xtab} employs a self-attention mechanism and an advanced tokenizer specifically designed for tabular data. This approach has been extended to combine multimodal tabular and echocardiographic inputs in \citep{painchaud2024fusingechocardiographyimagesmedical}.
Recently, symmetric cross-attention has been explored in IRENE \citep{Zhou2023}, enabling each modality to be sequentially contextualized by the other, as it is classically done in vision and language models~\citep{tan2019lxmert}. 
Although these multimodal fusion methods show improvements over the FT-Transformer trained on tabular data alone, they process the different modalities symmetrically, assuming equal relevance between them. This assumption does not align with the inherent asymmetry of our problem, where tabular data is the primary modality, and echocardiographic time series provide complementary but secondary information.
Finally, an alignment loss inspired by CLIP \citep{radfordLearningTransferableVisual2021a} has been used in MMCL \citep{Hager_2023_CVPR} to merge tabular data and medical images. While aligning tabular and echocardiographic representations is relevant in our context, applying a global alignment across all features from both modalities is overly restrictive, since tabular data contains information that is not present in echocardiographic videos.

\begin{figure}[h!]
    \centering
    \includegraphics[width=\linewidth]{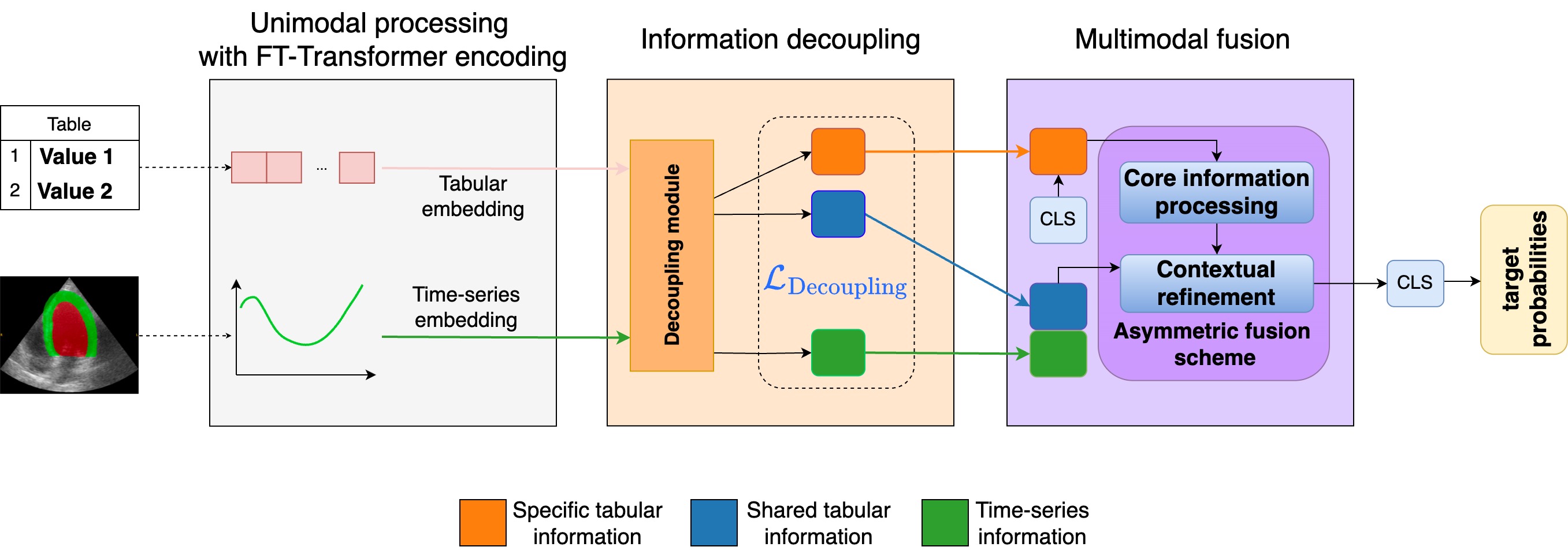}
    \caption{Overview of our model's global architecture. Tabular and echocardiographic data are first processed and tokenized separately. Each modality is encoded using a FT-Transformer \citep{FT-gorishny2021} before entering the decoupling module. The primary modality (tabular data) is then split into shared and modality-specific components. Finally, a multimodal fusion scheme is implemented, accounting for the asymmetric contribution of the two modalities. The primary modality drives core information processing, while the secondary modality provides contextual refinement.} 
    \label{fig:pipeline}
\end{figure}

This paper introduces a method for the characterization of cardiac hypertension that explicitly addresses multimodal fusion with asymmetric modalities, \ie, a primary source of information -- tabular data -- and a secondary source -- time series extracted from echocardiographic image sequences. The approach shown in Fig \ref{fig:pipeline} is devoted to effectively merge the primary and second multimodal inputs. Starting from a unimodal processing of each modality, we learn a relevant and structured latent space (middle in Fig \ref{fig:pipeline}) and introduce a fusion operator dedicated to asymmetric fusion (right in Fig \ref{fig:pipeline}). Our contributions can be summarized as follows:

\begin{itemize}
    \item We separate tabular data into specific information and information shared with echocardiographic time series. This is done using specialized loss functions that differentiate information types and use label supervision for self-regularization. This method enables better multimodal data integration by organizing the latent space into shared and modality-specific attributes.
   
    \item We introduce an asymmetric fusion scheme based on interleaved cross-attention, that prioritizes one modality while gradually contextualizing it with the secondary complementary information. 
    By emphasizing one modality and using the other for enhancement, we achieve a nuanced and effective integration of multimodal information.
\end{itemize}

\section{Methodology: Decoupling information and fusing asymmetric modalities}
In this section, we present the key contributions of the proposed method. First, a decoupling module re-expresses the primary modality (tabular data) into a modality-specific representation and a shared representation with the complementary modality (time series). \textcolor{black}{Given the asymmetry in information content between tabular data and echocardiographic videos, where the former may include details not visually apparent in the latter, it is reasonable to consider this multimodal data as inherently imbalanced in terms of information richness. Our approach is to decouple tabular data into two components: shared features, such as  left ventricular mass, which are represented in tabular data and time series, and specific tabular features, such as demographic attributes (e.g. age, sex), which are not present in time series. Our insight is to learn a space in which shared features are aligned, while keeping information from the specific and complementary information in tabular data.} 
Secondly, the rich tabular-specific representations are re-contextualized using the shared and time-series embeddings through an asymmetric fusion scheme. 
 
\subsection{Decoupling module}

In the multimodal framework, effectively integrating diverse data sources remains a critical challenge. A naive fusion may be suboptimal as one modality could be less informative. In our case, tabular data provide diverse and efficient information to characterize hypertension, while time series can offer additional valuable insights. To efficiently fuse these sources of information, we propose to decouple the tabular data into tabular-specific tokens and tokens that share common information with time series. To do so, an \textbf{information decoupling module} (see Figure \ref{fig:pipeline}) is introduced after the unimodal processing of the two modalities and before the fusion scheme.

\begin{figure}[t]
    \centering
    \includegraphics[width=0.8\linewidth]{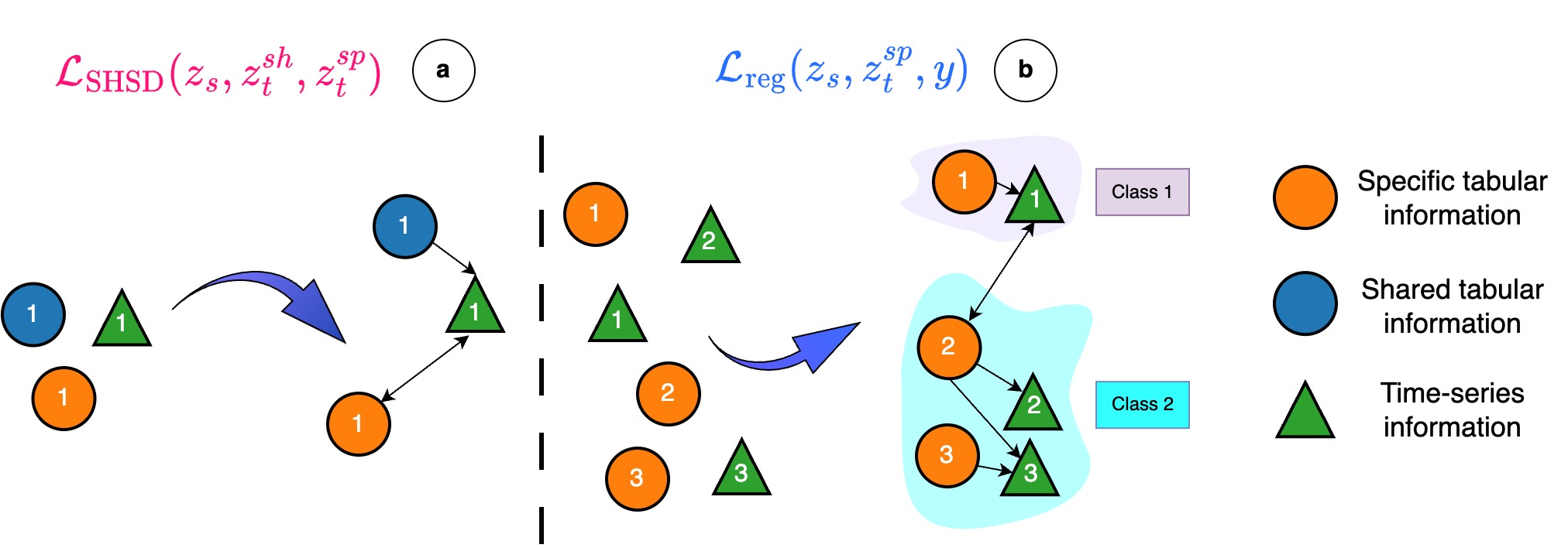}
    \caption{Decoupling module scheme in DAFTED. The SHSD loss enforces the tabular information separation by pushing the specific information away from time series while pulling the shared tabular information closer to it. Simultaneously, the regularization loss behaves similarly to the CLIP loss \citep{radfordLearningTransferableVisual2021a} but with label supervision \citep{NEURIPS2020_d89a66c7}. It specifically pulls together the specific tabular and time-series data belonging to the same class while pushing apart information from different classes} 
    \label{fig:decoupling-loss}
\end{figure}

Let denote $\bm t$ and $\bm s$ the embedded tabular data and time series respectively. The dataset consists of tabular data and time series for each patient, along with their corresponding labels $y$. We further introduce $g_\theta^t$ and $g_\phi^s$, the decoupling functions that project the modalities into modality-specific and shared latent spaces. We chose to use linear projections for the decoupling functions, as this approach demonstrated superior performance compared to alternative methods we evaluated. The resulting embeddings are defined as follows: $\bm z_s = g_\phi^s(\bm s)$ for time-series and $(\bm  z_t^{sp}, \bm  z_t^{sh}) = g_\theta^t(\bm t)$ for specific and shared tabular modality respectively. To enforce the information decoupling, we employ the following decoupling loss which enhances the alignment of shared information representations while simultaneously distinguishing modality-specific elements: 

\begin{align}
l_i^{s,t} = -\log\pars{\frac{\exp\{\mathrm{sim}(\zs{i}, \ztsh{i} )/\tau\}}{\sum_{k=1}^N\exp\{\mathrm{sim}({\zs{i},{\ztsp{k}}})/\tau\}}},
\label{eq:DCU}
\end{align}     

where $\mathrm{sim}(u, v) = u^\top v / ||u||\,||v||$ is the cosine similarity.
As illustrated in \ref{fig:decoupling-loss}-a, this contrastive loss uses the projected time-series representation vector $\bm{z}_s$ as the anchor point, toward which the representation $\bm{z}^{sh}_t$ is drawn, as they encapsulate similar information. Conversely, $\bm z_t^{sp}$ should be mapped far from the anchor point to preserve specific information from the primary modality. 
To maximize this contrastive effect, we aim to minimize the \textbf{SHared-Specific Decoupling} (SHSD) loss, defined as the sum of \ref{eq:DCU} and its symmetrical counterpart:

\begin{equation}    
\mathcal{L}_\mathrm{SHSD}(\zs{} , \ztsh{}, \ztsp{}) = \frac{1}{2N}\sum_{i=1}^N \pars{l_i^{s,t} + l_i^{t,s}}.
\end{equation}

Following the approach of \citet{decoupledcontrastive2022}, our method removes positive pairs from the denominator of the decoupling contrastive loss, addressing the \textit{negative-positive coupling} problem where positive samples in the denominator are inadvertently repelled. 

To reinforce the overall coherency of the latent space, we introduce a secondary loss that brings closer the representations sharing the same labels, while pushing apart the embeddings of other samples in the batch: 

\begin{equation}    
r^{t,s}_i  = - \frac{1}{S_i}\sum_{j=1}^N\mathds{1}{\{y_j = y_i\}}\log\left( \frac{\exp\braces{\mathrm{sim}\pars{ \ztsp{i} ,\, \zs{j} }/\tau}}{\sum_{k=1}^N \exp\braces{\mathrm{sim}\pars{ \ztsp{i} ,\, \zs{k} }/\tau}} \right),
\label{eq:SupCLIP}
\end{equation}
where $\mathds{1}{\{y_j = y_i\}}$ is an indicator function \textcolor{black}{and $S_i=\sum_{j=1}^N\mathds{1}{\{y_j = y_i\}}$}.

 The expected organization of the latent space following the minimization of this loss is illustrated in \ref{fig:decoupling-loss}-b. This loss acts as a regularization and ensures that specific tabular information $z_t^{sp}$ is positioned closer to the time-series representations $z_s$ of patients within the same label group.

Again, to equivalently update the model with respect to each modality, we use the sum of the loss \ref{eq:SupCLIP}  and its symmetrical. The regularization loss can thus be formulated as follows:
\begin{equation}
\mathcal{L}_\mathrm{reg}(\zs{}, \ztsp{}, y)  = \frac{1}{2N}\sum_{i=1}^N \pars{r^{t,s}_i + r^{s,t}_i}.
\end{equation}

Finally, the total decoupling loss denoted $\mathcal{L}_{\text{Decoupling}}$ is defined as the sum of the SHSD and regularization loss:
\begin{equation}
    \mathcal{L}_\mathrm{Decoupling}(\zs{}, \ztsp{}, \ztsh{}, y) = \mathcal{L}_\mathrm{SHSD}(\zs{}, \ztsh{}, \ztsp{}) + \mathcal{L}_\mathrm{reg}(\zs{}, \ztsp{}, y).
\end{equation}

\begin{figure}[t]
    \centering
    \includegraphics[width=0.8\linewidth]{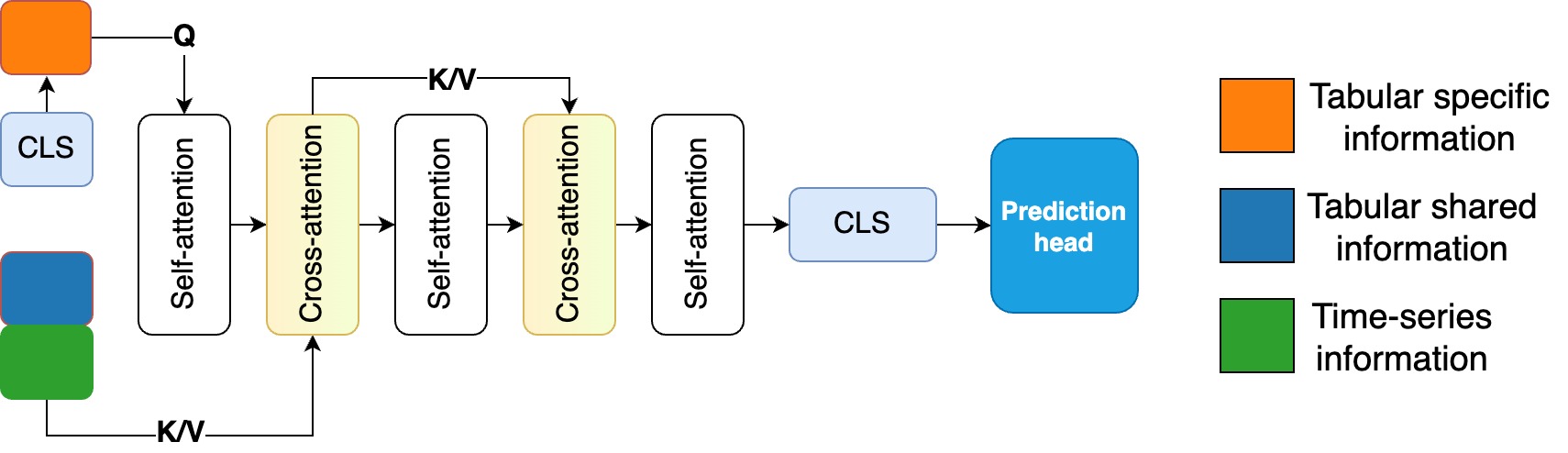}
    \caption{Asymmetric fusion scheme architecture. The tabular modality containing specific information is the prevailing modality that serves as the query tokens in all blocks, while the shared tabular and time-series information collaboratively enrich the primary modality. The cross-attention blocks employ shared weights, following \citet{jaeglePerceiverGeneralPerception2021a}, strategically enforcing an asymmetric relationship between modalities. The $\mathrm{[CLS]}$ token serves as the class aggregate representation for the final prediction head as in \citet{devlin-etal-2019-bert}}
    \label{fig:FT-Interleaved}
\end{figure}

\textcolor{black}{To improve the merging process, we propose in the following section an asymmetric fusion scheme that leverages the decoupling module to hierarchically integrate specific tabular information as the primary modality and shared tabular information with time series as sources of contextual refinement}

\subsection{Fusion scheme: interleaved attention modules}

Building on our earlier insight that modalities contribute asymmetrically to information content, we design a fusion scheme that prioritizes tabular data while using time series for contextual refinement. The Transformer architecture \citep{NIPS2017_3f5ee243-vaswani}, leveraging self- and cross-attention mechanisms, provides an efficient framework for processing asymmetrical information by treating the most informative modality as the query $\bm Q$, and the other as the context key $\bm K$ and value $\bm V$, as shown in Fig. \ref{fig:FT-Interleaved}.

In the fusion scheme, self-attention processes the primary tabular data, while cross-attention integrates context from shared tabular and time-series tokens. {Self-attention and cross-attention blocks alternate, where the cross-attention blocks use shared weights, establishing an asymmetry between the primary tabular data and the contextual refinement from time-series and shared tabular information.} A class token \citep{devlin-etal-2019-bert} is appended to the specific tabular tokens at the first stage of the architecture and serves as input to a prediction head at the final stage to perform classification. The entire pipeline is optimized by minimizing the final loss defined in Eq. \ref{eq:model_loss}:
\begin{equation}
\mathcal{L}(\hat{y}, y, \zs{}, \ztsh{}, \ztsp{}) = \mathcal{L}_\mathrm{CrossEntropy}(\hat{y},y) + \lambda \mathcal{L}_\mathrm{Decoupling}(\zs, \ztsp{},\ztsh{}, y),
\label{eq:model_loss}
\end{equation}
where $\lambda$ is a unique hyperparameter that balances the cross-entropy and decoupling terms.

\section{Experimental setup}

\paragraph{Dataset}CARDINAL is a valuable dataset combining echocardiographic image sequences from A2C and A4C views with comprehensive tabular data including demographics, lab results, and clinical exam measurements.
\citep{CARDINAL2023}. This multimodal data was collected on 239 patients at the Hospices Civils de Lyon, France, with the approval of the local ethics committee. The tabular data corresponds to 64 numerical and categorical descriptors, extracted from the EHR server. We used the hyper-tension severity (HT-severity) descriptor as the target to predict for each patient. It consists of three labels: \textit{wht} (White Coat Hypertension), for subjects with no positive diagnosis of hypertension; \textit{controlled}, for patients where the hypertension is managed to meet recommended blood pressure levels; and \textit{uncontrolled}, for patients who remain above these levels. Additional information regarding the data used in our study is provided in Appendix \ref{sec:data-info}.
\paragraph{Implementation Details} Training was conducted for 1000 epochs for DAFTED and all baseline models, the model exhibiting the lowest validation loss being selected for evaluation on the test dataset. All results utilized a temperature of $\tau = 0.1$ for the decoupling losses. We employed the Adam optimizer \citep{DBLP:journals/corr/KingmaB14} with a batch size of 128. The models were trained on 171 samples, evaluated on 20 samples, and tested on 48 samples, with a balanced distribution of the three target labels across these subsets. For the tabular modality, we selected 13 statistically independent features that are highly correlated with HT severity and easily measurable and accessible (unlike 24-hour measurements), while for the echocardiographic data, we retained the 7 measurements per view proposed by \citet{painchaud2024fusingechocardiographyimagesmedical}, resulting in 14 times-series automatically extracted from the two apical views using the segmentation framework described in \citep{CARDINAL2023}.  The FT-Transformer was configured to follow the XTab model by \citet{zhu2023xtab}, optimized for tabular data processing and featuring 3 transformer blocks with 8 heads of Self-Attention (MSA) and an embedding size of 192. To ensure fair comparisons, we maintained a uniform embedding size across all models. Given the class imbalance, we used ROC AUC metrics (one-vs-rest, averaged across the three target labels) for comparison.

\section{Results}

\paragraph{Baselines} The performance of our model was compared with XGBoost, a leading algorithm for tabular data analysis, in both single-modality and multimodal scenarios. For unimodal cases, we also present results of the FT-Transformer \citep{FT-gorishny2021} trained separately on tabular and time-series data. Among unimodal models, FT-Transformer was trained separately on tabular (Tab) and time-series (TS) data. XGBoost is characterized by its complexity rather than a specific number of parameters, making its size irrelevant for comparison with other models. We evaluate several state-of-the-art models that integrate tabular and imaging clinical data within a multimodal fusion framework, namely \citet{Hager_2023_CVPR}, \citet{Zhou2023}, and \citet{painchaud2024fusingechocardiographyimagesmedical}, alongside a naive fusion baseline that concatenates the modalities before feeding them into a two-layer perceptron. 
 
\begin{table}[h!]
\centering
\resizebox{0.48\textwidth}{!}{%
\begin{minipage}{0.5\textwidth}  
\centering
\floatconts{tab:results1}%
    {\caption{Comparison of our method with SOTA models. Mean and std are computed across 10 training runs with different seeds. FT-T refers to the FT-Transformer\bracketcite{FT-gorishny2021}.}}%
    {\begin{tabular}{l|cc}
    \toprule
    Model & $\text{ROC AUC}$ & \# parameters \\
    \midrule
    \multicolumn{3}{c}{Unimodal models}\\ 
    \midrule
    \addlinespace
    XGBoost \bracketcite{Chen2016XGboost} &$87.4$& N/A \\
    FT-T Tab &$85.8 \pm 4.8$& 863K \\
    FT-T TS &$52.2\pm 2.3$& 1.0M \\
    \midrule
    \multicolumn{3}{c}{Multimodal fusion models}\\
    \midrule
    XGBoost &$79.7$& N/A \\
    MLP &$81.8 \pm1.3$& 391K\\
    MMCL \bracketcite{Hager_2023_CVPR} &$77.4 \pm 2.1$& 1.6M  \\ 
    IRENE \bracketcite{Zhou2023} &$86.7 \pm 2.8$& 102.9M \\
    FT-T \bracketcite{painchaud2024fusingechocardiographyimagesmedical}&$88.9 \pm 1.4$& 1.1M\\
    \midrule
    \addlinespace
    \bfseries DAFTED (ours) & $\mathbf{91.0 \pm 0.7}$& 3.0M\\
    \bottomrule
    \end{tabular}}
\end{minipage}%
}
\hspace{0.056\textwidth} 
\resizebox{0.44\textwidth}{!}{
\begin{minipage}{0.5\textwidth}  
\centering
\small
\floatconts{tab:ablations}%
    {\caption{Impact of the decoupling module (Decoupling) and the asymmetric fusion scheme (Asym. fusion)
    }}%
    {\begin{tabular}{cc|c}
   
    Decoupling&  Asym. fusion &$\text{ROC AUC}$ \\
    \toprule
    \addlinespace
    \ding{55}&\ding{55}&$89.4\pm 2.2$\\
    \addlinespace
    \textcolor{black}{\ding{55}}&\textcolor{black}{\ding{51}}& \textcolor{black}{$90.4\pm 1.0$} \\
    \addlinespace 
   
   \ding{51}&\ding{51} & $\mathbf{91.0 \pm 0.7}$\\
    \end{tabular}}
    \vspace{0.4cm}
    \centering
    \includegraphics[width=\linewidth]{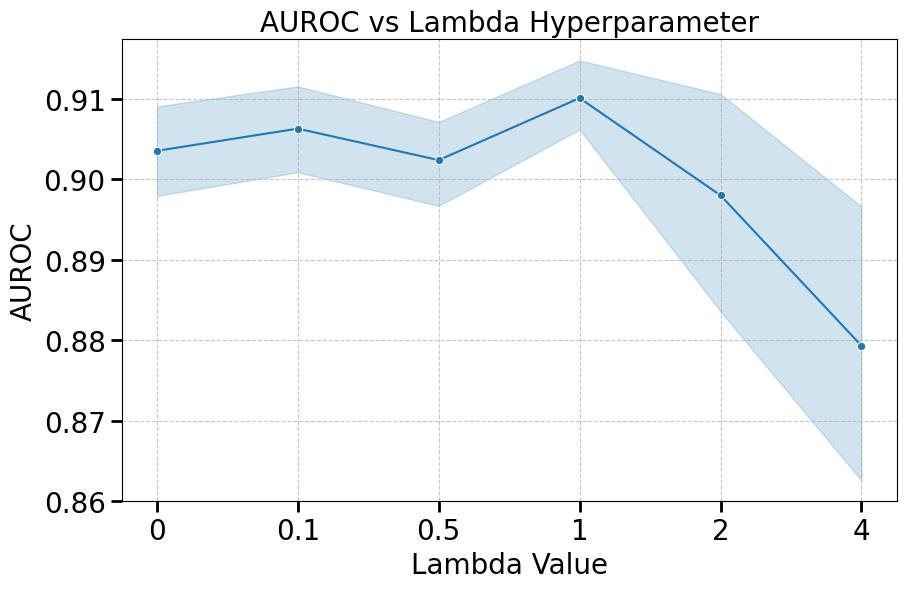}
    \captionof{figure}{Sensitivity of $\lambda$ in our loss function.}
    \label{fig:lmbda}
\end{minipage}
}
\end{table}

\paragraph{State-of-the-art comparison} Results are presented in Table \ref{tab:results1}. XGBoost remains a strong competitor, demonstrating the best results in unimodal scenarios, while time-series data alone lacks consistent predictive power. Naive concatenation of tabular and time-series data in MLP and XGBoost models underperforms compared to unimodal tabular models or more advanced fusion approaches. Our method excels by prioritizing tabular data complemented by echocardiographic time series with a ROC AUC more than 2pt superior to top models equally weighting both modalities. Our results were achieved without complex measurements like 24-hour systolic and diastolic blood pressure, demonstrating effective performance using simpler data acquisition methods. The paired t-tests in Appendix \ref{sec:ttest} show that our method significantly outperforms state-of-the-art baselines, with a 1\% significance threshold for each comparison.

\paragraph{Ablations and model analysis} 

\begin{figure}[h!]
    \centering
    \resizebox{\textwidth}{!}{%
    \begin{minipage}{0.47\textwidth}
        \centering
        \includegraphics[width=\linewidth]{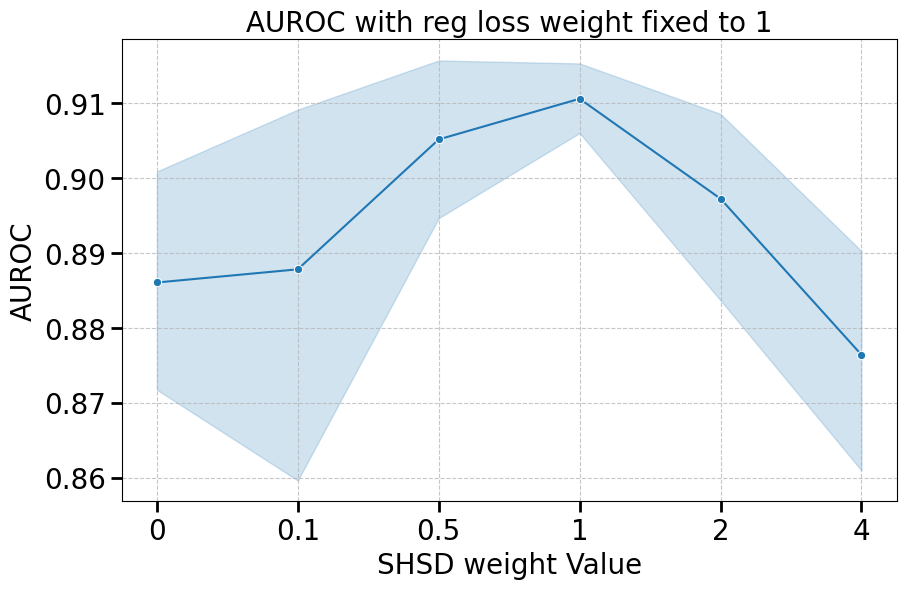}
        \caption{\textcolor{black}{Sensitivity of the $\mathrm{SHSD}$ loss weight}}
        \label{fig:shsd-sensi}
    \end{minipage}
    \hspace{0.09cm}
        \begin{minipage}{0.47\textwidth}
        \centering
        \includegraphics[width=\linewidth]{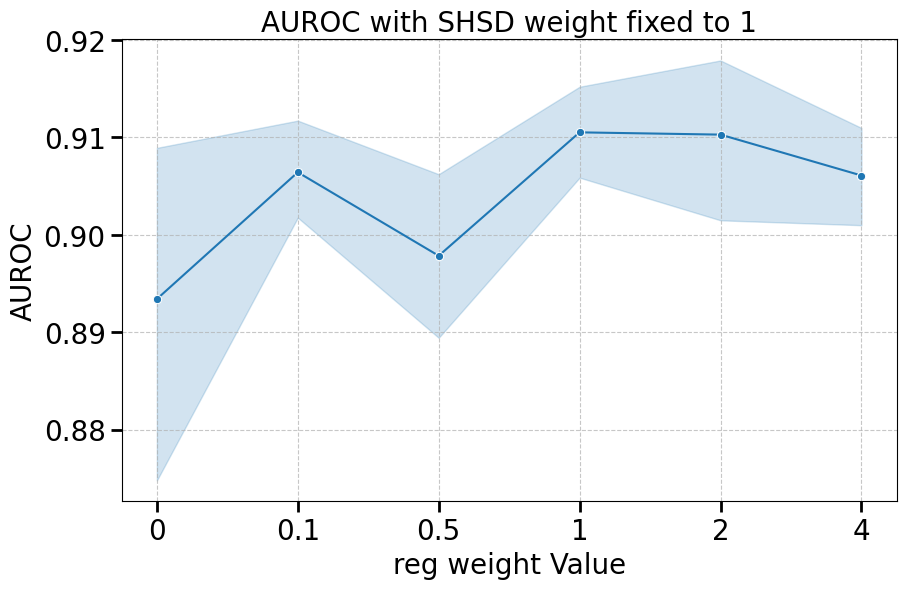}
        \caption{\textcolor{black}{Sensitivity of the $\mathrm{reg}$ loss weight}}
        \label{fig:reg-sensi}
    \end{minipage}
    }
\end{figure}

\begin{figure}[h!]
    \centering
    \begin{minipage}{0.49\textwidth}
        \centering
        \includegraphics[width=\linewidth]{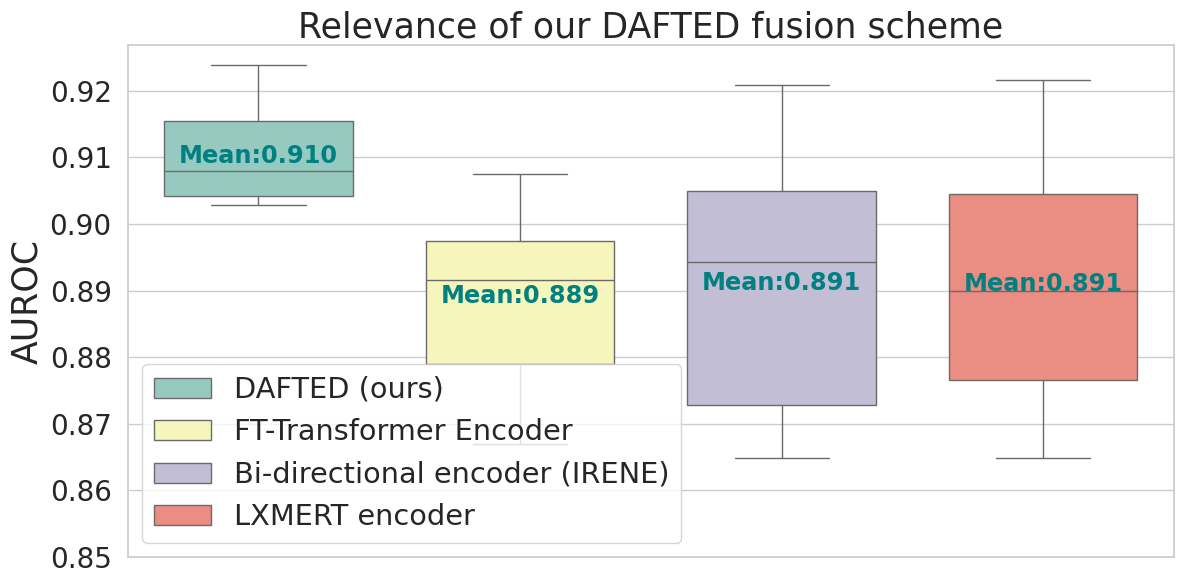}
        \caption{Impact of DAFTED fusion module vs SOTA fusion schemes}
        \label{fig:boxplotfusion}
    \end{minipage}
    \hspace{0.06cm}
    \begin{minipage}{0.49\textwidth}
        \centering
        \includegraphics[width=\linewidth]{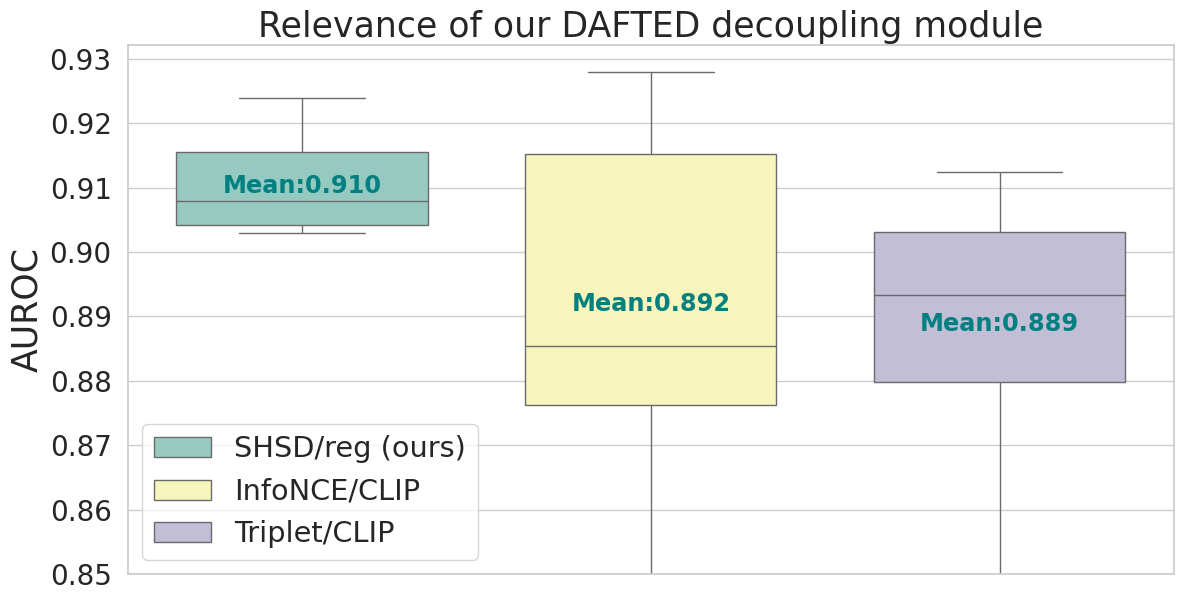}
        \caption{Impact of DAFTED decoupling module vs contrastive baselines}
        \label{fig:boxplotloss}
    \end{minipage}
    \label{fig:boxplot-figures}
\end{figure}

We conducted ablation studies to evaluate the contribution of the decoupling loss and the asymmetric fusion scheme.
Results presented in Table \ref{tab:ablations} show that our decoupled asymmetric fusion scheme significantly increases the performance of diagnosis by more than 2pt. We also evaluated various fusion schemes and contrastive losses within our framework. Fig. \ref{fig:boxplotfusion} compares alternative fusion schemes, while Fig. \ref{fig:boxplotloss} explores different decoupling loss mechanisms. Classical approaches, such as InfoNCE \citep{sohnImprovedDeepMetric2016} and Triplet loss \citep{weinbergerDistanceMetricLearning} fail to match the performance of our proposed fusion scheme, highlighting the innovative nature of our multimodal representation learning strategy. We report paired t-tests in Appendix \ref{sec:ttest} to quantitatively assess the significance of our method's performance. Fig. \ref{fig:lmbda} investigates the influence of the hyperparameter $\lambda$ in our loss function. Results show the robustness of our method, with an optimal value of $\lambda=1$. \textcolor{black}{Fig. \ref{fig:shsd-sensi} and \ref{fig:reg-sensi} further investigate the sensitivity of our method to the weights of the SHSD and regularization losses. This process involved setting one weight to a value of 1, while varying the other weight across a range of values from 0 to 4. Interestingly, this approach inherently includes an ablation study when one weight is set to 0 while the other remains at 1. The results demonstrate the synergistic effect of the two components of the decoupling loss. In particular, it shows the robustness of the decoupling loss across a broad range of the regularization weights, from 1 to substantially higher values. In contrast, the SHSD loss appears to be most effective in guiding the model to its optimal performance when set at a value of 1. This suggests that while the regularization component of our approach is flexible, the SHSD loss plays a crucial role in fine-tuning the model's capabilities, with its impact being most pronounced at this specific weighting. Finally, these results confirm that both components are essential for the decoupling module to be effective, with the best outcomes achieved when the weights are set to 1.} 


\begin{figure}[h!]
    \centering
    \resizebox{\textwidth}{!}{%
    \begin{minipage}{0.47\textwidth}
        \centering
        \includegraphics[width=\linewidth]{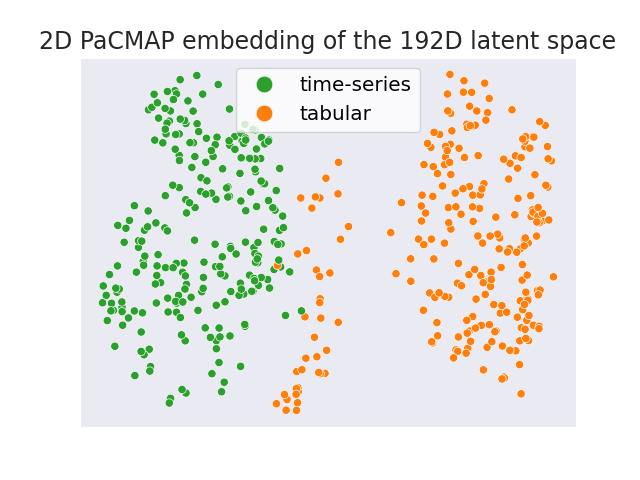}
        \caption{PacMAP representation of latent vectors before fusion without decoupling}
        \label{fig:pacmad_nodecoup}
    \end{minipage}
    \label{fig:pacmap-figures}
    \hspace{0.09cm}
        \begin{minipage}{0.47\textwidth}
        \centering
        \includegraphics[width=\linewidth]{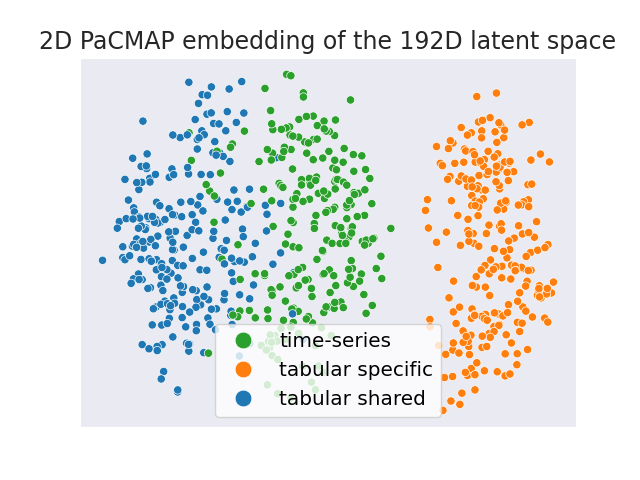}
        \caption{PacMAP representation of latent vectors before fusion with decoupling}
        \label{fig:pacmap_dafted}
    \end{minipage}
    }
\end{figure}

\paragraph{Latent space structuration} To visualize the structure of the latent space, we reduce the dimension of the latent vectors of each modality (tabular specific, tabular shared and time series) with the PacMAP method \citep{JMLR:v22:20-1061} before fusion. We observe that without decoupling the tabular information, tabular and time-series modalities are slightly intertwined (Fig. \ref{fig:pacmad_nodecoup}). Furthermore, Fig. \ref{fig:pacmap_dafted} shows that the decoupling loss forces the shared tabular information to be closer to the time-series representations, while specific tabular embeddings are well separated from the time series.

\section{Conclusion}

In this study, we propose a new fusion strategy based on an asymmetric scheme. A first decoupling module separates the primary tabular modality into tokens specific to this modality and tokens shared with the secondary modality. A dedicated fusion scheme is then employed to integrate the secondary modality as a contextual refinement. Results show significant improvement compared to SOTA models, as well as stability across different parameters and architectural configurations. This study demonstrates that the characterization of HT severity can be achieved using a limited amount of easily measurable patient data, which can ultimately improve patient care.

\clearpage  
\midlacknowledgments{We would like to thank Elisa Le Maout (ARC, Hôpital Lyon Sud, Hospices Civils de Lyon, Lyon, France) for her help with data collection for the CARDINAL dataset. This research is conducted within the ORCHID project, which receives funding from the French National Research Agency (ANR) (ANR-22-CE45-0029-01). For the purpose of open access, the authors have applied a CC BY public copyright license to any Author Accepted
Manuscript (AAM) version arising from this submission. We acknowledge the financial support provided by PEPR Sharp (ANR-23-PEIA-0008, ANR,
FRANCE 2030). }

\bibliography{midl25-14}

\begin{thebibliography}{24}
\providecommand{\natexlab}[1]{#1}
\providecommand{\url}[1]{\texttt{#1}}
\expandafter\ifx\csname urlstyle\endcsname\relax
  \providecommand{\doi}[1]{doi: #1}\else
  \providecommand{\doi}{doi: \begingroup \urlstyle{rm}\Url}\fi

\bibitem[Chen and Guestrin(2016)]{Chen2016XGboost}
Tianqi Chen and Carlos Guestrin.
\newblock Xgboost: A scalable tree boosting system.
\newblock In \emph{Proceedings of the 22nd ACM SIGKDD International Conference
  on Knowledge Discovery and Data Mining}, pages 785--794. ACM, 2016.
\newblock \doi{10.1145/2939672.2939785}.
\newblock URL \url{https://dl.acm.org/doi/10.1145/2939672.2939785}.

\bibitem[Devlin et~al.(2019)Devlin, Chang, Lee, and
  Toutanova]{devlin-etal-2019-bert}
Jacob Devlin, Ming-Wei Chang, Kenton Lee, and Kristina Toutanova.
\newblock {BERT}: Pre-training of deep bidirectional transformers for language
  understanding.
\newblock In Jill Burstein, Christy Doran, and Thamar Solorio, editors,
  \emph{Proceedings of the 2019 Conference of the North {A}merican Chapter of
  the Association for Computational Linguistics: Human Language Technologies,
  Volume 1 (Long and Short Papers)}, pages 4171--4186, Minneapolis, Minnesota,
  June 2019. Association for Computational Linguistics.
\newblock \doi{10.18653/v1/N19-1423}.
\newblock URL \url{https://aclanthology.org/N19-1423}.

\bibitem[Gorishniy et~al.(2021)Gorishniy, Rubachev, Khrulkov, and
  Babenko]{FT-gorishny2021}
Yury Gorishniy, Ivan Rubachev, Valentin Khrulkov, and Artem Babenko.
\newblock Revisiting deep learning models for tabular data.
\newblock In M.~Ranzato, A.~Beygelzimer, Y.~Dauphin, P.S. Liang, and J.~Wortman
  Vaughan, editors, \emph{Advances in Neural Information Processing Systems},
  volume~34, pages 18932--18943. Curran Associates, Inc., 2021.
\newblock URL
  \url{https://proceedings.neurips.cc/paper_files/paper/2021/file/9d86d83f925f2149e9edb0ac3b49229c-Paper.pdf}.

\bibitem[Grinsztajn et~al.(2022)Grinsztajn, Oyallon, and
  Varoquaux]{grinsztajn2022why}
Leo Grinsztajn, Edouard Oyallon, and Gael Varoquaux.
\newblock Why do tree-based models still outperform deep learning on typical
  tabular data?
\newblock In \emph{Thirty-sixth Conference on Neural Information Processing
  Systems Datasets and Benchmarks Track}, 2022.
\newblock URL \url{https://openreview.net/forum?id=Fp7__phQszn}.

\bibitem[Hager et~al.(2023)Hager, Menten, and Rueckert]{Hager_2023_CVPR}
Paul Hager, Martin~J. Menten, and Daniel Rueckert.
\newblock Best of both worlds: Multimodal contrastive learning with tabular and
  imaging data.
\newblock In \emph{Proceedings of the IEEE/CVF Conference on Computer Vision
  and Pattern Recognition (CVPR)}, pages 23924--23935, June 2023.

\bibitem[Jaegle et~al.(2021)Jaegle, Gimeno, Brock, Vinyals, Zisserman, and
  Carreira]{jaeglePerceiverGeneralPerception2021a}
Andrew Jaegle, Felix Gimeno, Andy Brock, Oriol Vinyals, Andrew Zisserman, and
  Joao Carreira.
\newblock Perceiver: {{General Perception}} with {{Iterative Attention}}.
\newblock In \emph{Proceedings of the 38th {{International Conference}} on
  {{Machine Learning}}}, pages 4651--4664. PMLR, July 2021.

\bibitem[Khosla et~al.(2020)Khosla, Teterwak, Wang, Sarna, Tian, Isola,
  Maschinot, Liu, and Krishnan]{NEURIPS2020_d89a66c7}
Prannay Khosla, Piotr Teterwak, Chen Wang, Aaron Sarna, Yonglong Tian, Phillip
  Isola, Aaron Maschinot, Ce~Liu, and Dilip Krishnan.
\newblock Supervised contrastive learning.
\newblock In H.~Larochelle, M.~Ranzato, R.~Hadsell, M.F. Balcan, and H.~Lin,
  editors, \emph{Advances in Neural Information Processing Systems}, volume~33,
  pages 18661--18673. Curran Associates, Inc., 2020.
\newblock URL
  \url{https://proceedings.neurips.cc/paper_files/paper/2020/file/d89a66c7c80a29b1bdbab0f2a1a94af8-Paper.pdf}.

\bibitem[Kingma and Ba(2015)]{DBLP:journals/corr/KingmaB14}
Diederik~P. Kingma and Jimmy Ba.
\newblock Adam: {A} method for stochastic optimization.
\newblock In Yoshua Bengio and Yann LeCun, editors, \emph{3rd International
  Conference on Learning Representations, {ICLR} 2015, San Diego, CA, USA, May
  7-9, 2015, Conference Track Proceedings}, 2015.
\newblock URL \url{http://arxiv.org/abs/1412.6980}.

\bibitem[Ling et~al.(2023)Ling, Painchaud, Courand, Jodoin, Garcia, and
  Bernard]{CARDINAL2023}
Hang~Jung Ling, Nathan Painchaud, Pierre-Yves Courand, Pierre-Marc Jodoin,
  Damien Garcia, and Olivier Bernard.
\newblock Extraction of volumetric indices from echocardiography: Which deep
  learning solution for clinical use?
\newblock In Olivier Bernard, Patrick Clarysse, Nicolas Duchateau, Jacques
  Ohayon, and Magalie Viallon, editors, \emph{Functional Imaging and Modeling
  of the Heart}, pages 245--254, Cham, 2023. Springer Nature Switzerland.
\newblock ISBN 978-3-031-35302-4.

\bibitem[Mancia et~al.(2023)Mancia, Kreutz, Brunström, Burnier, Grassi,
  Januszewicz, Muiesan, Tsioufis, Agabiti-Rosei, Algharably, Azizi, Benetos,
  Borghi, Hitij, Cifkova, Coca, Cornelissen, Cruickshank, Cunha, Danser, Pinho,
  Delles, Dominiczak, Dorobantu, Doumas, Fernández-Alfonso, Halimi, Járai,
  Jelaković, Jordan, Kuznetsova, Laurent, Lovic, Lurbe, Mahfoud, Manolis,
  Miglinas, Narkiewicz, Niiranen, Palatini, Parati, Pathak, Persu, Polonia,
  Redon, Sarafidis, Schmieder, Spronck, Stabouli, Stergiou, Taddei,
  Thomopoulos, Tomaszewski, Van~de Borne, Wanner, Weber, Williams, Zhang, and
  Kjeldsen]{mancia_2023_2023}
Giuseppe Mancia, Reinhold Kreutz, Mattias Brunström, Michel Burnier, Guido
  Grassi, Andrzej Januszewicz, Maria~Lorenza Muiesan, Konstantinos Tsioufis,
  Enrico Agabiti-Rosei, Engi Abd~Elhady Algharably, Michel Azizi, Athanase
  Benetos, Claudio Borghi, Jana~Brguljan Hitij, Renata Cifkova, Antonio Coca,
  Veronique Cornelissen, J.~Kennedy Cruickshank, Pedro~G. Cunha, A.~H.~Jan
  Danser, Rosa Maria~de Pinho, Christian Delles, Anna~F. Dominiczak, Maria
  Dorobantu, Michalis Doumas, María~S. Fernández-Alfonso, Jean-Michel Halimi,
  Zoltán Járai, Bojan Jelaković, Jens Jordan, Tatiana Kuznetsova, Stephane
  Laurent, Dragan Lovic, Empar Lurbe, Felix Mahfoud, Athanasios Manolis, Marius
  Miglinas, Krzystof Narkiewicz, Teemu Niiranen, Paolo Palatini, Gianfranco
  Parati, Atul Pathak, Alexandre Persu, Jorge Polonia, Josep Redon, Pantelis
  Sarafidis, Roland Schmieder, Bart Spronck, Stella Stabouli, George Stergiou,
  Stefano Taddei, Costas Thomopoulos, Maciej Tomaszewski, Philippe Van~de
  Borne, Christoph Wanner, Thomas Weber, Bryan Williams, Zhen-Yu Zhang, and
  Sverre~E. Kjeldsen.
\newblock 2023 {ESH} {Guidelines} for the management of arterial hypertension
  {The} {Task} {Force} for the management of arterial hypertension of the
  {European} {Society} of {Hypertension}.
\newblock \emph{J. Hypertens.}, 41:\penalty0 1874, December 2023.
\newblock ISSN 0263-6352.
\newblock \doi{10.1097/HJH.0000000000003480}.
\newblock URL
  \url{https://journals.lww.com/jhypertension/fulltext/2023/12000/2023_esh_guidelines_for_the_management_of_arterial.2.aspx}.

\bibitem[Painchaud et~al.(2024)Painchaud, Stym-Popper, Courand, Thome, Jodoin,
  Duchateau, and Bernard]{painchaud2024fusingechocardiographyimagesmedical}
Nathan Painchaud, Jérémie Stym-Popper, Pierre-Yves Courand, Nicolas Thome,
  Pierre-Marc Jodoin, Nicolas Duchateau, and Olivier Bernard.
\newblock Fusing echocardiography images and medical records for continuous
  patient stratification, 2024.
\newblock URL \url{https://arxiv.org/abs/2401.07796}.

\bibitem[Radford et~al.(2021)Radford, Kim, Hallacy, Ramesh, Goh, Agarwal,
  Sastry, Askell, Mishkin, Clark, Krueger, and
  Sutskever]{radfordLearningTransferableVisual2021a}
Alec Radford, Jong~Wook Kim, Chris Hallacy, Aditya Ramesh, Gabriel Goh,
  Sandhini Agarwal, Girish Sastry, Amanda Askell, Pamela Mishkin, Jack Clark,
  Gretchen Krueger, and Ilya Sutskever.
\newblock Learning {{Transferable Visual Models From Natural Language
  Supervision}}.
\newblock In \emph{Proceedings of the 38th {{International Conference}} on
  {{Machine Learning}}}, pages 8748--8763. PMLR, July 2021.

\bibitem[Sohn(2016)]{sohnImprovedDeepMetric2016}
Kihyuk Sohn.
\newblock Improved {{Deep Metric Learning}} with {{Multi-class N-pair Loss
  Objective}}.
\newblock In \emph{Advances in {{Neural Information Processing Systems}}},
  volume~29. Curran Associates, Inc., 2016.

\bibitem[Tan and Bansal(2019)]{tan2019lxmert}
Hao Tan and Mohit Bansal.
\newblock Lxmert: Learning cross-modality encoder representations from
  transformers.
\newblock In \emph{Proceedings of the 2019 Conference on Empirical Methods in
  Natural Language Processing}, 2019.

\bibitem[Tsehay et~al.(2017)Tsehay, Lay, Roth, Wang, Kwak, Turkbey, Pinto,
  Wood, and Summers]{10.1117/12.2254423}
Yohannes~K. Tsehay, Nathan~S. Lay, Holger~R. Roth, Xiaosong Wang, Jin~Tae Kwak,
  Baris~I. Turkbey, Peter~A. Pinto, Brad~J. Wood, and Ronald~M. Summers.
\newblock {Convolutional neural network based deep-learning architecture for
  prostate cancer detection on multiparametric magnetic resonance images}.
\newblock In Samuel G.~Armato III and Nicholas~A. Petrick, editors,
  \emph{Medical Imaging 2017: Computer-Aided Diagnosis}, volume 10134, page
  1013405. International Society for Optics and Photonics, SPIE, 2017.
\newblock \doi{10.1117/12.2254423}.
\newblock URL \url{https://doi.org/10.1117/12.2254423}.

\bibitem[Vaswani et~al.(2017)Vaswani, Shazeer, Parmar, Uszkoreit, Jones, Gomez,
  Kaiser, and Polosukhin]{NIPS2017_3f5ee243-vaswani}
Ashish Vaswani, Noam Shazeer, Niki Parmar, Jakob Uszkoreit, Llion Jones,
  Aidan~N Gomez, \L~ukasz Kaiser, and Illia Polosukhin.
\newblock Attention is all you need.
\newblock In I.~Guyon, U.~Von Luxburg, S.~Bengio, H.~Wallach, R.~Fergus,
  S.~Vishwanathan, and R.~Garnett, editors, \emph{Advances in Neural
  Information Processing Systems}, volume~30. Curran Associates, Inc., 2017.
\newblock URL
  \url{https://proceedings.neurips.cc/paper_files/paper/2017/file/3f5ee243547dee91fbd053c1c4a845aa-Paper.pdf}.

\bibitem[Wang et~al.(2021)Wang, Huang, Rudin, and Shaposhnik]{JMLR:v22:20-1061}
Yingfan Wang, Haiyang Huang, Cynthia Rudin, and Yaron Shaposhnik.
\newblock Understanding how dimension reduction tools work: An empirical
  approach to deciphering t-sne, umap, trimap, and pacmap for data
  visualization.
\newblock \emph{Journal of Machine Learning Research}, 22\penalty0
  (201):\penalty0 1--73, 2021.
\newblock URL \url{http://jmlr.org/papers/v22/20-1061.html}.

\bibitem[Wang et~al.(2024)Wang, Gao, Xiao, and
  Sun]{wangMediTabScalingMedical2024a}
Zifeng Wang, Chufan Gao, Cao Xiao, and Jimeng Sun.
\newblock {{MediTab}}: {{Scaling Medical Tabular Data Predictors}} via {{Data
  Consolidation}}, {{Enrichment}}, and {{Refinement}}.
\newblock In \emph{Proceedings of the {{Thirty-ThirdInternational Joint
  Conference}} on {{Artificial Intelligence}}}, pages 6062--6070, Jeju, South
  Korea, August 2024. International Joint Conferences on Artificial
  Intelligence Organization.
\newblock ISBN 978-1-956792-04-1.
\newblock \doi{10.24963/ijcai.2024/670}.

\bibitem[Weinberger and Saul(2009)]{weinbergerDistanceMetricLearning}
Kilian~Q Weinberger and Lawrence~K Saul.
\newblock Distance {{Metric Learning}} for {{Large Margin Nearest Neighbor
  Classification}}.
\newblock \emph{JMLR}, 2009.

\bibitem[Xu et~al.(2023)Xu, Zhu, and Clifton]{xu_multimodal_2023}
Peng Xu, Xiatian Zhu, and David~A. Clifton.
\newblock Multimodal {Learning} {With} {Transformers}: {A} {Survey}.
\newblock \emph{IEEE PAMI}, 45:\penalty0 12113--32, October 2023.
\newblock ISSN 1939-3539.
\newblock \doi{10.1109/TPAMI.2023.3275156}.
\newblock URL \url{https://ieeexplore.ieee.org/abstract/document/10123038}.

\bibitem[Yeh et~al.(2022)Yeh, Hong, Hsu, Liu, Chen, and
  LeCun]{decoupledcontrastive2022}
Chun-Hsiao Yeh, Cheng-Yao Hong, Yen-Chi Hsu, Tyng-Luh Liu, Yubei Chen, and Yann
  LeCun.
\newblock Decoupled contrastive learning.
\newblock In \emph{Computer Vision – ECCV 2022: 17th European Conference, Tel
  Aviv, Israel, October 23–27, 2022, Proceedings, Part XXVI}, page 668–684,
  Berlin, Heidelberg, 2022. Springer-Verlag.
\newblock ISBN 978-3-031-19808-3.
\newblock \doi{10.1007/978-3-031-19809-0_38}.
\newblock URL \url{https://doi.org/10.1007/978-3-031-19809-0_38}.

\bibitem[Yi et~al.(2022)Yi, Ou, Hu, Qiu, Quan, Othmane, Wang, and
  Wu]{10.3389/fphys.2022.918381}
Zhenglin Yi, Zhenyu Ou, Jiao Hu, Dongxu Qiu, Chao Quan, Belaydi Othmane,
  Yongjie Wang, and Longxiang Wu.
\newblock Computer-aided diagnosis of prostate cancer based on deep neural
  networks from multi-parametric magnetic resonance imaging.
\newblock \emph{Frontiers in Physiology}, 13, 2022.
\newblock ISSN 1664-042X.
\newblock \doi{10.3389/fphys.2022.918381}.
\newblock URL
  \url{https://www.frontiersin.org/journals/physiology/articles/10.3389/fphys.2022.918381}.

\bibitem[Zhou et~al.(2023)Zhou, Yu, Wang, et~al.]{Zhou2023}
H.Y. Zhou, Y.~Yu, C.~Wang, et~al.
\newblock A transformer-based representation-learning model with unified
  processing of multimodal input for clinical diagnostics.
\newblock \emph{Nature Biomedical Engineering}, 7:\penalty0 743--755, 2023.
\newblock \doi{10.1038/s41551-023-01045-x}.

\bibitem[Zhu et~al.(2023)Zhu, Shi, Erickson, Li, Karypis, and
  Shoaran]{zhu2023xtab}
Bingzhao Zhu, Xingjian Shi, Nick Erickson, Mu~Li, George Karypis, and Mahsa
  Shoaran.
\newblock Xtab: Cross-table pretraining for tabular transformers.
\newblock In \emph{Proceedings of the International Conference on Machine
  Learning (ICML)}, pages 181--204, 2023.

\end{thebibliography}

\newpage

\appendix

\section{Hyperparameter sensitivity test}

\begin{figure}[hbtp]
    \centering
    \includegraphics[width=0.7\linewidth]{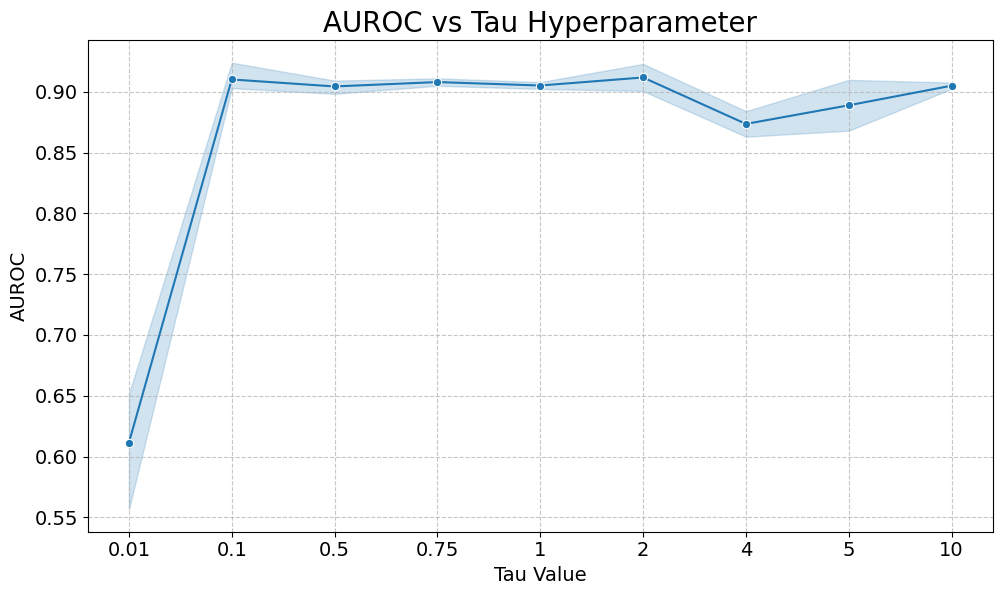}
    \caption{Hyperparameter $\tau$ sensitivity test, results averaged across 10 seeds}
    \label{fig:tau}
\end{figure}
Figure \ref{fig:tau} presents our analysis of the hyperparameter $\tau$, associated with the contrastive losses in equations \eqref{eq:DCU} and \eqref{eq:SupCLIP}. We observe that performance metrics remain relatively stable for $\tau$ values above 0.1, demonstrating our model's robustness to temperature scaling.

\begin{figure}[hbtp]
    \centering
    \includegraphics[width=0.9\linewidth]{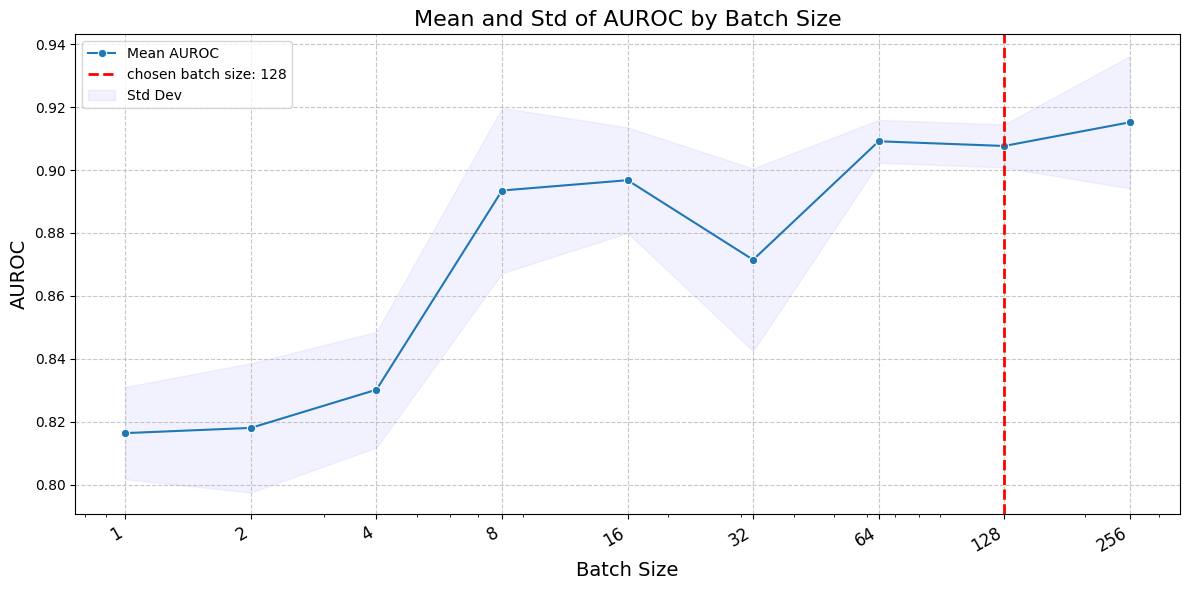}
    \caption{Batch size sensitivity test, results averaged across 10 seeds}
    \label{fig:bs-test}
\end{figure}

\section{Statistical Tests}
\label{sec:ttest}
\paragraph{Paired-samples t-Test} In Table \ref{tab:ttest_results}, we conducted multiple paired t-tests on related samples to evaluate whether our proposed model, DAFTED, significantly outperforms the other baseline models and alternative approaches. These statistical analyses aim to assess the comparative effectiveness of DAFTED against existing methodologies in the field.

\begin{table}[h!]
\centering
\caption{Related-paired t-test Results Comparing DAFTED ROC metrics against baselines}
\begin{tabular}{l|ccc}
\textbf{Models} & \textbf{Mean Difference} & \textbf{t-statistic} & \textbf{p-value} \\
\toprule
\multicolumn{4}{c}{Relevance of our method against SOTA baselines}\\
\midrule
\addlinespace
FT-Transformer & $2.05$ & $4.62$ & $0.0014$  \\
IRENE &$4.31$ & $4.53$ & $0.0014$ \\
MLP &$9.22$ & $23.88$&$\ll0.001$ \\ 
MMCL & $ 13.57$&$24.01$ & $\ll0.001$\\
\addlinespace
\midrule
\multicolumn{4}{c}{Impact of our decoupling asymmetric fusion scheme (DAF)}\\
\midrule
\addlinespace
Our method without DAF &$2.11$&$3.46$&$0.0071$\\
\bottomrule
\end{tabular}
\label{tab:ttest-baselines}
\end{table}

\begin{table}[h!]
\centering
\captionsetup{width=0.8\linewidth}
\caption{Related-paired t-test Results Comparing DAFTED ROC metrics against alternative decoupling modules and fusion schemes}

\resizebox{0.8\textwidth}{!}{
\begin{tabular}{l|ccc}
\textbf{Models} & \textbf{Mean Difference} & \textbf{t-statistic} & \textbf{p-value} \\
\toprule
\multicolumn{4}{c}{Alternative fusion modules (with decoupling)}\\
\midrule
\addlinespace
FT-Transformer & $1.59$&$3.22$&$0.0104$\\
{Bi-directional (IRENE)} & $1.92$&$2.66$&$0.0259$\\
LXMERT &$1.93$&$2.87$&$ 0.0184$\\
\addlinespace
\midrule

\multicolumn{4}{c}{Alternative decoupling modules (with our DAFTED fusion scheme)}\\
\midrule
\addlinespace

InfoNCE/CLIP &$ 1.84$&$ 1.93$& $0.0856$\\
Triplet/CLIP &$2.15$&$3.14$& $0.0119$\\
\bottomrule
\end{tabular}
}
\label{tab:ttest_results}
\end{table}

\newpage
\section{Additional results}
\begin{table}[hbtp]  
\centering
\floatconts{tab:additional-res}%
    {\caption{Other metrics to compare our model DAFTED againt baselines and SOTA models}}%
    {\begin{tabular}{l|ccc}
    \toprule
    &AUROC&AUPRC&F1-score \\
    \midrule
    XGBoost &$87.4$&$77.6$& $65.4$\\
    MLP &$81.8 \pm 1.3$&$59.8 \pm 2.0$ & $52.2 \pm 6.8$ \\
    FT-Transformer &$88.9 \pm 1.1$&$79.6 \pm 2.7$& $68.7 \pm 3.6$ \\
    \midrule
     \textbf{DAFTED (ours)} &$ \bm{91.0 \pm 0.7}$& $\bm{82.2 \pm 4.8}$& $\bm{69.9 \pm 5.1}$ \\
    \bottomrule
    \end{tabular}}
\end{table}%
\clearpage
\section{Loss details}
\begin{table}[h!]
\centering
\captionsetup{width=0.8\linewidth}
\caption{Loss details}

\resizebox{\textwidth}{!}{
\begin{tabular}{l|ll}
\textbf{Name} & \textbf{Symbol} & \textbf{Equation}  \\
\toprule
Cross-entropy & $\mathcal{L}_{\text{CrossEntropy}}(y,\hat{y})$&$\sum_{i=1}^N y_i \log(\hat{y}_i)$\\
\addlinespace
\addlinespace
Shared-Specific Decoupling (SHSD) & $\mathcal{L}_\mathrm{SHSD}(\zs{}, \ztsh{}, \ztsp{})$&$l_i^{s,t} = -\log\pars{\frac{\exp\{\mathrm{sim}(\zs{i}, \ztsh{i} )/\tau\}}{\sum_{k=1}^N\exp\{\mathrm{sim}({\zs{i},{\ztsp{k}}})/\tau\}}}$\\
\addlinespace
\addlinespace
Regularization (reg) &$\mathcal{L}_\mathrm{reg}(\zs{}, \ztsp{}, y)$&$r^{t,s}_i  = - \frac{1}{S_i}\sum_{j=1}^N\mathds{1}{\{y_j = y_i\}}\log\left( \frac{\exp\braces{\mathrm{sim}\pars{ \ztsp{i} ,\, \zs{j} }/\tau}}{\sum_{k=1}^N \exp\braces{\mathrm{sim}\pars{ \ztsp{i} ,\, \zs{k} }/\tau}} \right)$\\
\bottomrule
\end{tabular}
}
\label{tab:loss-details}
\end{table}

\section{Hyperparameters}
We detail here the hyperparameter choices for training both our model and the baselines. We experimented with various values for the decoupling loss weight, temperature scale, and batch size, selecting the configuration that yielded the best performance for our model. The number of layers in the transformer unimodal encoders was kept fixed to align with state-of-the-art architectures that have demonstrated strong results across multiple tasks. For each hyperparameter, we specify the final value used in our model training.

\begin{table}[h!]
\centering
\caption{Hyperparameter details}

\begin{tabular}{c|cc}
Hyperparams. & Role & Value \\
\toprule
$\lambda$ & \makecell{Balances the decoupling loss, adjusting its weight\\ comparing the classification cross-entropy loss} &  $\lambda = 1$\\
\midrule
$\tau$ & \makecell{Denotes the temperature scale parameter \\ for the contrastive decoupling losses}& $\tau=0.1$\\
\midrule
$\mathrm{batch~size}$ & Number of samples processed per batch in the dataset & $\mathrm{bs} = 128$ \\
\bottomrule
\end{tabular}
\label{tab:hyperparameter-details}
\end{table}

\newpage
\section{Data specification}
\label{sec:data-info}
\setlength{\arrayrulewidth}{0.5mm}
\setlength{\tabcolsep}{18pt}
\renewcommand{\arraystretch}{1.1}
\newcolumntype{s}{>{\columncolor[HTML]{A9A9A9}} m{3cm}}
\begin{table}[h!]
\arrayrulecolor[HTML]{A9A9A9}
\centering
\rowcolors{2}{white}{gray!25}
\footnotesize
\footnotesize\caption{List of 13 patient descriptors for the CARDINAL dataset extracted from Electronic Health Records (EHRs)}
\begin{tabular}{ s|m{1.8cm}|m{5cm}}
 \rowcolor{white}Abbreviation& Unit/Labels &Description \\
\hline
age  &years & Age \\
sbp\_tte & mmHg & Systolic Blood Pressure (SBP) during TTE \\
pp\_tte & mmHg& Pulse Pressure (SBP) during TTE  \\
diastolic\_dysfunction  &0--4 & {1 point per parameter of diastolic dysfunction:  \textit{dilated\_la}, \textit{reduced\_e\_prime}, \textit{d\_dysfunction\_ratio}, \textit{ph\_vmax\_tr}} \\
pw\_d & cm & Left ventricular Posterior Wall (PW) thickness at end-Diastole (D) \\
lvm\_ind & g/m\textsuperscript{2} & Left Ventricular Mass (LVM) indexed to BSA    \\
e\_e\_prime\_ratio & -- & Ratio of E velocity over e': E/e' \\
gfr  & mL/min/1.73m\textsuperscript{2} & Glomerular Filtration Rate (GFR) indexed to standard body surface area \\
lateral\_e\_prime & cm/s & Lateral mitral annular velocity (e') \\
septal\_e\_prime & cm/s & Septal mitral annular velocity (e') \\
a\_velocity & m/s & A-wave (active blood flow caused by atrial contraction) velocity \\
ddd & -- & Defined Daily Dose (DDD) of blood pressure medication \\
la\_volume & mL/m\textsuperscript{2} & Left Atrial (LA) volume indexed to body surface area (BSA) \\
\hline
\end{tabular}
\end{table}
\begin{table}[h!]
\arrayrulecolor[HTML]{A9A9A9}
\centering
\rowcolors{2}{white}{gray!25}
\footnotesize 
\footnotesize\caption{List of 7 patient descriptors extracted from segmentations of transthoracic echocardiogram (TTE) for the CARDINAL dataset, extracted frame-by-frame, available for apical 4 chamber (A4C) and apical 2 chamber (A2C) views}
\begin{tabular}{ s|m{1.8cm}|m{5cm}}

 \rowcolor{white}Abbreviation& Unit/Labels &Description \\
\hline
 lv\_area & cm\textsuperscript{2} & Surface area of the LV \\
 lv\_length & cm  & Distance between the LV's apex and midpoint at the base \\
 gls & \% & Global Longitudinal Strain (GLS) \\
 ls\_left & \% & Regional Longitudinal Strain (LS) at the base of the left wall \textit{A4C left wall}: septum / \textit{A2C left wall}: inferior\\
 ls\_right & \%  & Regional Longitudinal Strain (LS) at the base of the right wall \textit{A4C right wall}: lateral / \textit{A2C right wall}: anterior \\
 myo\_thickness\_left & cm  & Average myocardial thickness at the base of the left wall \textit{A4C left wall}: septum / \textit{A2C left wall}: inferior \\
 myo\_thickness\_right & cm & Average myocardial thickness at the base of the right wall \textit{A4C right wall}: lateral / \textit{A2C right wall}: anterior\\
\hline
\end{tabular}
\end{table}
\end{document}